\documentclass{article}

\usepackage{times}
\usepackage{graphicx} 
\usepackage{subfigure}
\usepackage{amsmath}
\usepackage{amssymb}

\usepackage{algorithm}
\usepackage[noend]{algpseudocode}

\usepackage{natbib}
\usepackage{varwidth}
\usepackage{subfigure}

\usepackage{hyperref}

\usepackage[accepted]{icml2014} 

\DeclareMathOperator*{\argmin}{arg\,min}

\icmltitlerunning{Contrastive Structured Anomaly Detection for Gaussian Graphical Models}

\begin{document} 

\twocolumn[
\icmltitle{Contrastive Structured Anomaly Detection for Gaussian Graphical Models}

\icmlauthor{Abhinav Maurya}{amaurya@andrew.cmu.edu}
\icmladdress{Department of Information Systems, Carnegie Mellon University,
            Pittsburgh, PA -- 15213}
\icmlauthor{Mark Cheung}{markcheu@andrew.cmu.edu}
\icmladdress{Department of Electrical and Computer Engineering,
            Carnegie Mellon University, Pittsburgh PA -- 15213}

\icmlkeywords{Anomaly Detection, Gaussian Graphical Model, ADMM}

\vskip 0.3in
]

\begin{abstract}
Gaussian graphical models (GGMs) are probabilistic tools of choice for analyzing conditional dependencies between variables in complex systems. Finding changepoints in the structural evolution of a GGM is therefore essential to detecting anomalies in the underlying system modeled by the GGM. In order to detect structural anomalies in a GGM, we consider the problem of estimating changes in the precision matrix of the corresponding Gaussian distribution. We take a two-step approach to solving this problem:- (i) estimating a background precision matrix using system observations from the past without any anomalies, and (ii) estimating a foreground precision matrix using a sliding temporal window during anomaly monitoring. Our primary contribution is in estimating the foreground precision using a novel contrastive inverse covariance estimation procedure. In order to accurately learn only the structural changes to the GGM, we maximize a penalized log-likelihood where the penalty is the $l_1$ norm of difference between the foreground precision being estimated and the already learned background precision. We modify the alternating direction method of multipliers (ADMM) algorithm for sparse inverse covariance estimation to perform contrastive estimation of the foreground precision matrix. Our results on simulated GGM data show significant improvement in precision and recall for detecting structural changes to the GGM, compared to a non-contrastive sliding window baseline.

\end{abstract}

\section{Introduction}
Gaussian graphical models (GGMs) have been widely applied as probabilistic models for analyzing conditional dependencies between variables in complex systems such as gene regulatory networks \cite{dobra2004sparse} and flight planning systems \cite{liu2013learning}. The structure of a GGM is encoded in the inverse covariance matrix, also known as the precision matrix. The zero entries of the precision matrix correspond to pairs of features that are conditionally independent given the rest of the graph. Thus, estimation of the precision matrix can be thought of as estimating the topology of a Gaussian MRF \cite{koller2009probabilistic}. This estimation can be performed using several methods such as selection procedure \cite{scheinberg2010sparse} and graphical lasso \cite{yuan2007model}. Given the feature observations, a estimated sparse precision matrix can be used to uncover relationships between the features.

However, GGMs model stationary processes and temporal extensions to GGM focus on modeling the smoothly evolving nature of dynamic processes, not on anomaly detection. An example of such temporal modeling is \cite{kolar2010estimating} which employs a sliding window approach based on a semi-parametric class of models to estimate the graph structure locally. Real-time analysis for structural anomaly detection in GGMs is useful in a wide variety of applications such as detecting organizational process disruption, detecting vulnerabilities, and modeling gene regulation anomalies that cause diseases.

Another approach to anomaly detection would be to treat anomaly detection as a two-class modeling problem using joint graphical lasso \cite{danaher2014joint,yuan2007model,hoefling2010path}. The background and foreground precision matrices could then be simultaneously estimated using a joint convex optimization problem. However, if the structural change is localized or its effect is subtle, a penalized log-likelihood maximization using the joint graphical lasso will capture the background structure in both the estimated precision matrices and treat the structural change as statistical noise to best explain the observed dataset.

\section{Contrastive Structured Anomaly Detection}

We detect sudden structural changes that may occur in Gaussian graphical models (GGMs). In the \emph{data modeling phase}, we use background data from the past that has been identified to not contain any anomalies, to learn a GGM that describes the structural relationships between the random variables in the background data-generating process. In the \emph{anomaly monitoring phase}, we move a sliding window over newly arriving data and perform detection within each window. Given a set of new observations of the random variables within a window i.e. foreground datapoints, we intend to learn the minimum structural changes in the background graphical model that can explain the new set of observations. Hence, we call our algorithm \emph{Contrastive Structured Anomaly Detection (CSAD)} for Gaussian graphical model.

We can accomplish our goal by solving the graphical lasso optimization problem via penalizing the resulting foreground graphical model to be as close as possible in structure to the background graphical model. For very high values of penalization, we expect to see a foreground graphical model that is structurally identical to the background graphical model. As we reduce the penalization, we expect to see the most statistically significant structural changes to the foreground graphical model that explain the resultant foreground data. This method of finding structural changes can be advantageous when we have huge amounts of background data but the number of foreground observations is small compared to the dimensionality of the graphical model, and we expect localized structural changes to occur in the graphical model in response to an event.

Since the graphical lasso optimization is a semi-definite program, we solve the resulting optimization problem in a scalable fashion by using a distributed optimization algorithm called alternating direction method of multipliers (ADMM).

\subsection{Optimization Problem}
Using background data, we estimate a GGM precision matrix $\Theta_b$ that describes the structural relationships between the covariates \cite{boyd2011distributed}. Given a set of new observations, we can learn the structural changes by contrasting it with the background precision matrix via penalized log likelihood maximization:

\begin{equation}
\argmin_{\Theta \succeq 0} \{  \textrm{trace}(S\Theta) - \log\textrm{det}(\Theta) + \lambda ||\Theta - \Theta_b||_1 \}
\label{objective_eq}
\end{equation}

Here, $S$ is the empirical covariance matrix, $\Theta_b$ is the background precision matrix and $\Theta$ is the foreground precision matrix that we want to learn to detect structural changes in the underlying graphical model. For small $n$, the optimization problem is feasible since the MLE is tractable.

The augmented Lagrangian of the problem is as follows:

\begin{equation}
\begin{split}
L_{\rho} (\Theta, Z, U) & =  \textrm{trace}(S \Theta) - \log\textrm{det}(\Theta)  \\
&\quad   + \lambda || Z-\Theta_b ||_1 \\
&\quad  + \frac{\rho}{2} ||\Theta-Z+U||^2_F
\end{split}
\end{equation}

where \textrm{$\rho$} is the penalty parameter for the inequality between $\Theta$ and $Z$, \textrm{$\lambda$} is the lasso regularization parameter, and \textrm{$||*||_F$} is the Frobenius norm, the square root of the sum of squares of a matrix's entries.

\subsection{Optimization Algorithm}

We solve the optimization problem by alternatively optimizing over $\Theta$, $Z$, and $U$ \cite{boyd2011distributed}.

We can minimize the objective over \(\Theta\) using the first-order optimality condition for the gradient:

\begin{equation}
S-\Theta^{-1}+ \rho (\Theta-Z^n+U^n)=0
\end{equation}

\begin{equation}
\therefore \rho \Theta -\Theta^{-1}= \rho (Z^n-U^n)-S
\label{e1}
\end{equation}

Taking the eigenvalue decomposition of $\rho (Z^n-U^n)-S$ as $Q \Lambda Q^T$:

\begin{equation}
\rho \Theta -\Theta^{-1} = Q \Lambda Q^T
\label{e2}
\end{equation}

\begin{equation}
\therefore \rho \tilde{\Theta} -\tilde{\Theta}^{-1}= \Lambda
\end{equation}

where \(\Theta = Q \tilde{\Theta} Q^T \)

We solve for a diagonal solution of \(\tilde{\Theta}\) using the quadratic formula:

\begin{equation}
\tilde{\Theta}_{ii} = \frac{\lambda_i+\sqrt{\lambda_i^2+4\rho}}{2\rho}
\end{equation}

where $\lambda_i$ is the $i^{th}$ diagonal value of $\Lambda$. The computational cost of updating \(\Theta\) is thus an eigenvalue decomposition of a symmetric matrix.

The  minimization involving \textrm{$Z$} can be accomplished using elementwise soft-thresholding:

\begin{equation}
Z^{n+1}_{ij} = S_{\frac{\lambda}{\rho}}(\Theta^{n+1}_{ij} +U^{n+1}_{ij}-{\Theta_b}_{ij})
\end{equation}

Finally, we update $U$ to its new value as follows:

\begin{equation}
U^{n+1} = \Theta^{n+1}- Z^{n+1}+ U^{n}
\end{equation}

We summarize the ADMM procedure in algorithm (\ref{admm-algorithm}).

\begin{algorithm}[h]
\caption{ADMM Algorithm for CSAD}
\begin{algorithmic}
\Procedure{ADMM}{$S, \Theta_b, \lambda,\rho$}
\State $n = \textrm{size}(S)$ \\
\State $\Theta^1, Z^1, U^1 = \textrm{zeros}(n)$ \\
\For{$n=1..max\_iterations$} \\
\State \begin{varwidth}{\linewidth} $\Theta^{n+1} = \argmin_{\Theta} \textrm{trace}(S\Theta)$
\par \hskip \algorithmicindent $-\log\textrm{det} (\Theta) +\frac{\rho}{2}||\Theta-Z^n+U^n||^2_F$ \end{varwidth} \\
\State \begin{varwidth}{\linewidth} $Z^{n+1} = \argmin_{Z} \lambda ||Z-\Theta_b||_1$
\par \hskip \algorithmicindent $+\frac{\rho}{2}||\Theta^{n+1}-Z+U^n||^2_F$  \end{varwidth} \\
\State $U^{n+1} = \Theta^{n+1}- Z^{n+1}+ U^{n}$ \\
\EndFor
\State \textbf{return} $\Theta^{n+1}$
\EndProcedure
\end{algorithmic}
\label{admm-algorithm}
\end{algorithm}

\section{Experimental Setup and Results}

We simulated a 100-dimensional sparse positive semi-definite (PSD) precision matrix $P_b$ for background data. We also simulated a sparse PSD matrix $P_\delta$ of identical size for the anomalous change in precision matrix of background data. We obtain the foreground precision matrix $P_f$ by adding $P_b$ and $P_\delta$. Since $P_b$ and $P_\delta$ are both PSD matrices, their sum $P_f$ is also a PSD matrix since the space of PSD matrices is a convex cone.

Using the precision matrices $P_b$ and $P_f$ for background and foreground data, we simulate 10000 datapoints each to obtain the background and foreground datasets. We use zero vector as the mean of the multivariate Gaussian distributions being simulated.

\begin{figure}[h]
  \centering
    \includegraphics[trim=0mm 60mm 10mm 60mm,clip=true,width=\linewidth]{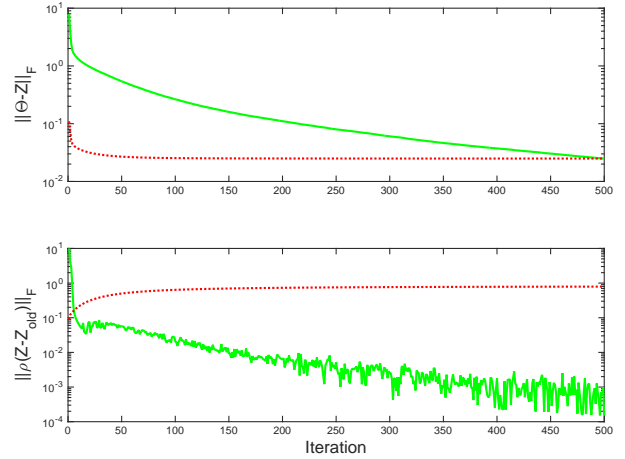}
    \caption{Convergence criteria versus iteration. The green solid curve shows the actual value, while the red dotted curve shows the tolerance for the convergence criterion. When both the values have fallen below their tolerances, the algorithm stops.}
    \label{fig:objective-values}
\end{figure}

\begin{figure}[h]
  \centering
    \includegraphics[trim=10mm 25mm 0mm 20mm,clip=true,width=0.42\textwidth]{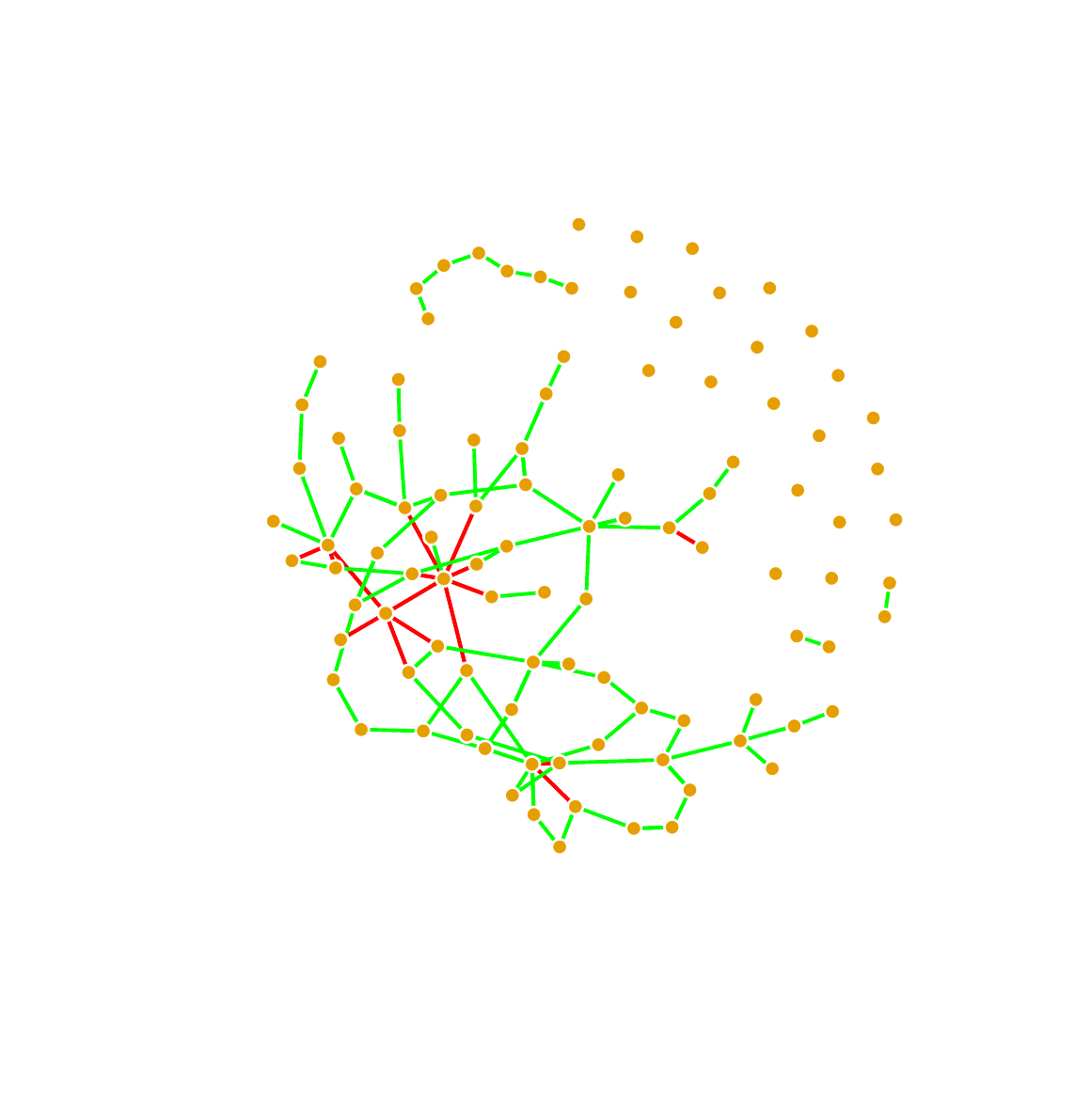}
    \caption{CSAD Detection. The edges correctly detected by CSAD are shown in green, while the edges incorrectly detected i.e. false positives are shown in red. All other edges are hidden for clarity.}
    \label{fig:csad-graph}
\end{figure}

\begin{figure}[h]
  \centering
    \includegraphics[trim=10mm 25mm 0mm 20mm,clip=true,width=0.42\textwidth]{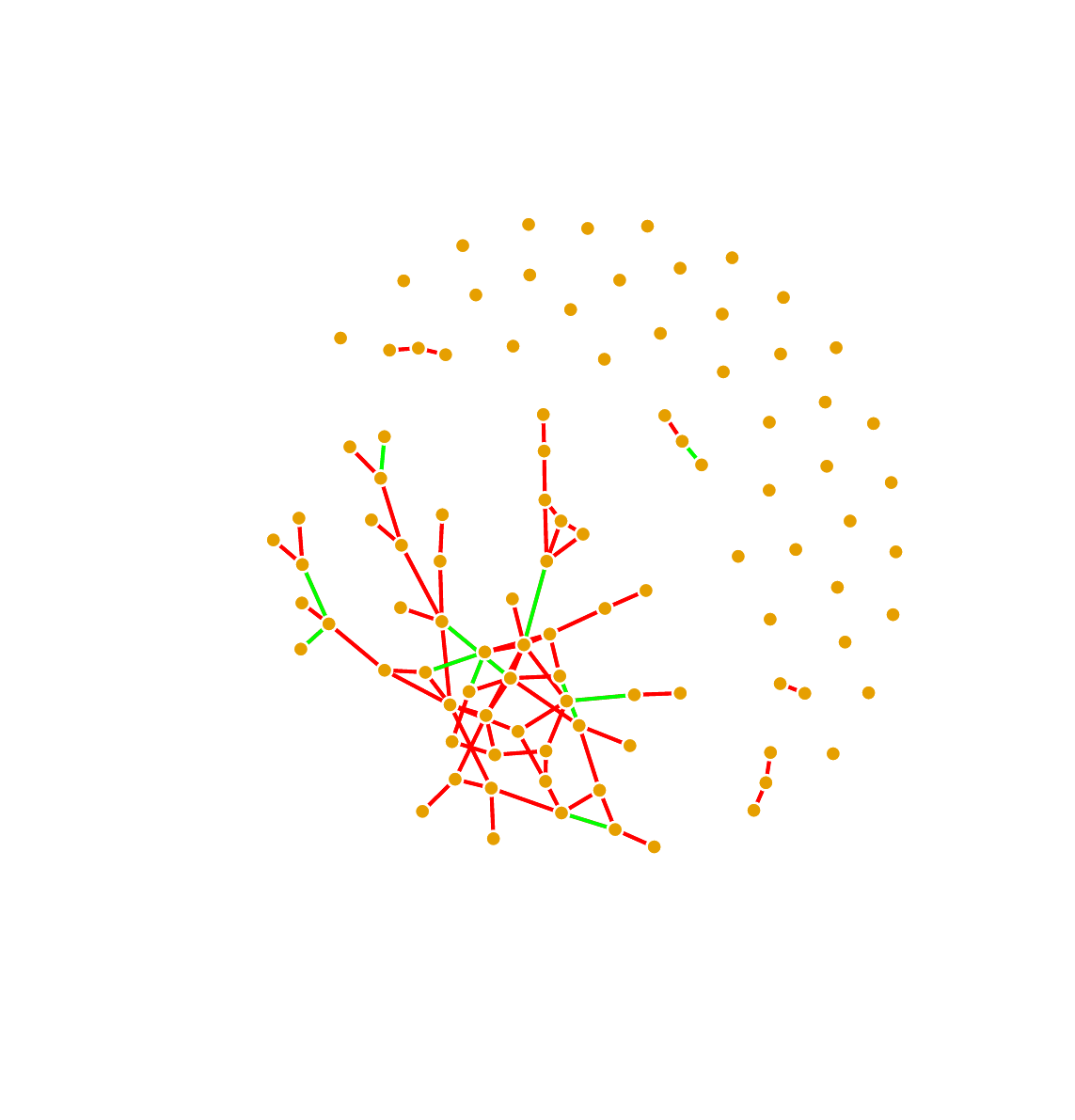}
    \caption{BSAD Detection. The true positive edges detected by BSAD are shown in green, while the false positives are shown in red. All other edges are hidden for clarity.}
    \label{fig:bsad-graph}
\end{figure}

Figure \ref{fig:objective-values} shows that the ADMM algorithm convergence is rapid. The ADMM algorithm is considered to have converged when the primal and dual residuals (green solid lines in figure \ref{fig:objective-values}) are small i.e. \(||\Theta-Z||_F \leq \epsilon^{primal}\) and \(||\rho(Z-Z_{old})||_F \leq \epsilon^{dual}\). The tolerance values $\epsilon^{primal}$ and $\epsilon^{dual}$ depend on the scale of the ADMM variables:

\begin{equation}
\epsilon^{primal} = n \epsilon^{abs}+\epsilon^{rel} \textrm{max}\{||\Theta||_F, ||Z||_F\}
\end{equation}

\begin{equation}
\epsilon^{dual} = n \epsilon^{abs}+\epsilon^{rel} ||\rho U||_F
\end{equation}

For our simulations, we chose \(\epsilon^{abs}=10^{-4}\) and \(\epsilon^{rel}=10^{-2}\), and the resulting convergence graph shows both the dual and primal residuals crossing the stopping criteria after around 500 iterations.

In order to compare our method with a baseline, we create a non-contrastive version of the structural anomaly detection algorithm which we call \emph{Baseline Structured Anomaly Detection (BSAD)}. In BSAD, we estimate the foreground precision matrix using sparse inverse covariance selection with only the foreground data i.e. background data or background precision matrix is not used. Such a method will work well if the structural change is strong, but will fare poorly if the change is localized and its strength is small.

For the precision matrices simulated to generate the data as well as the precision matrices estimated using algorithm BSAD or CSAD, we consider an edge exists between two nodes if the corresponding entry in the precision matrix is non-zero, and vice versa.

In figure (\ref{fig:csad-graph}), we show an example result of CSAD detection. The graph consists of 100 nodes. The edges correctly detected by CSAD i.e. the ones which were present in the injected structural change are shown in green. The edges which were falsely identified as part of the structural change but did not occur in the true injected change are shown in red. All other edges in hidden for reason of figure clarity. From the figure, we see that CSAD has a fairly high precision. Figure (\ref{fig:bsad-graph}) shows the same result for an example run of BSAD. We observe the significantly low precision offered by BSAD, since most edges are red.

Considering the presence of an edge in a structural change as 1 and its absence as 0. For example, an edge which was a part of the injected structural change but was not detected by the algorithm is considered a false negative.

\begin{figure}[h]
  \centering
    \includegraphics[page=1,width=0.9\linewidth]{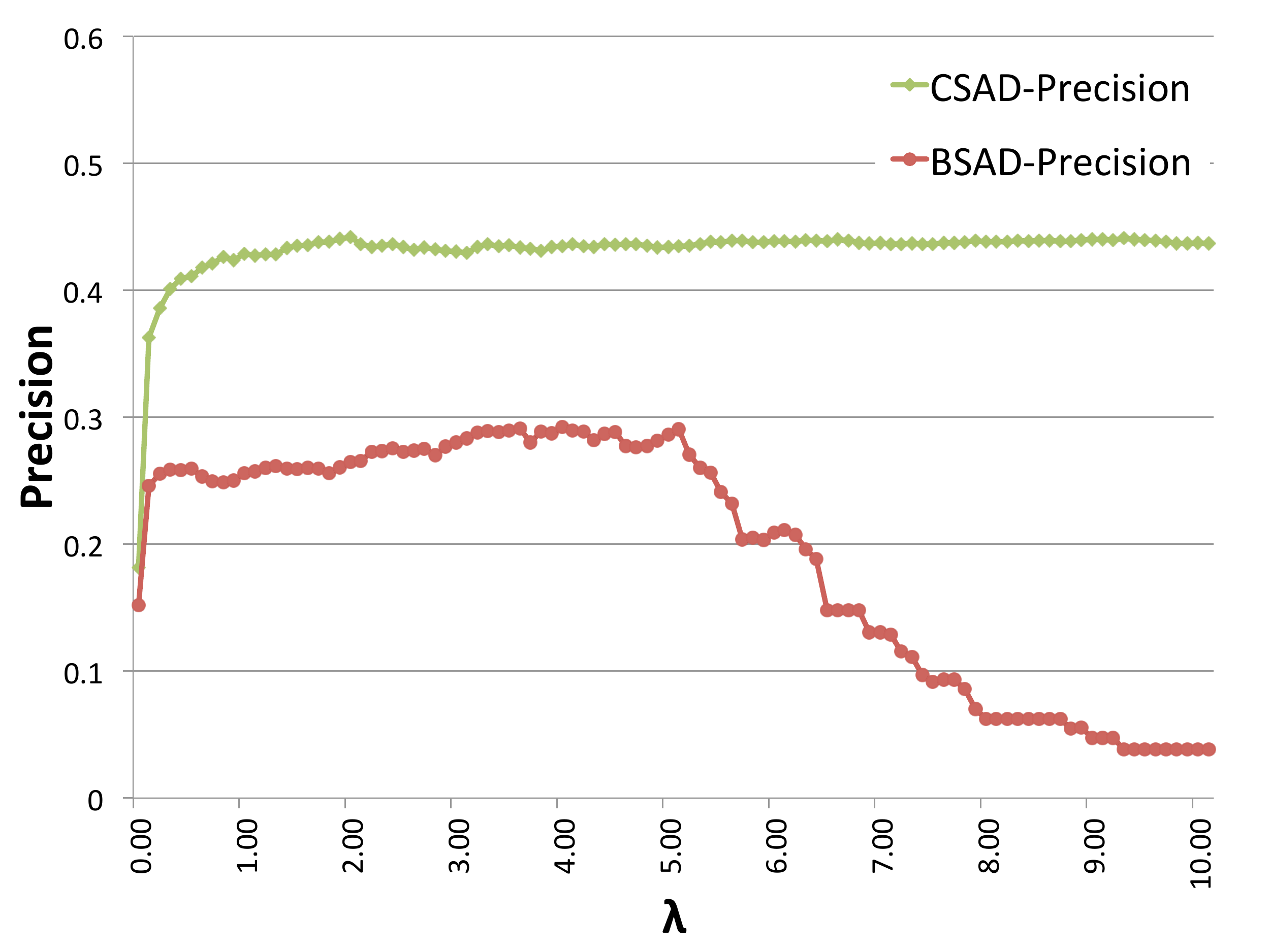}
    \caption{Precision of CSAD and BSAD}
    \label{fig:precision}
\end{figure}

\begin{figure}[h]
  \centering
    \includegraphics[page=2,width=0.9\linewidth]{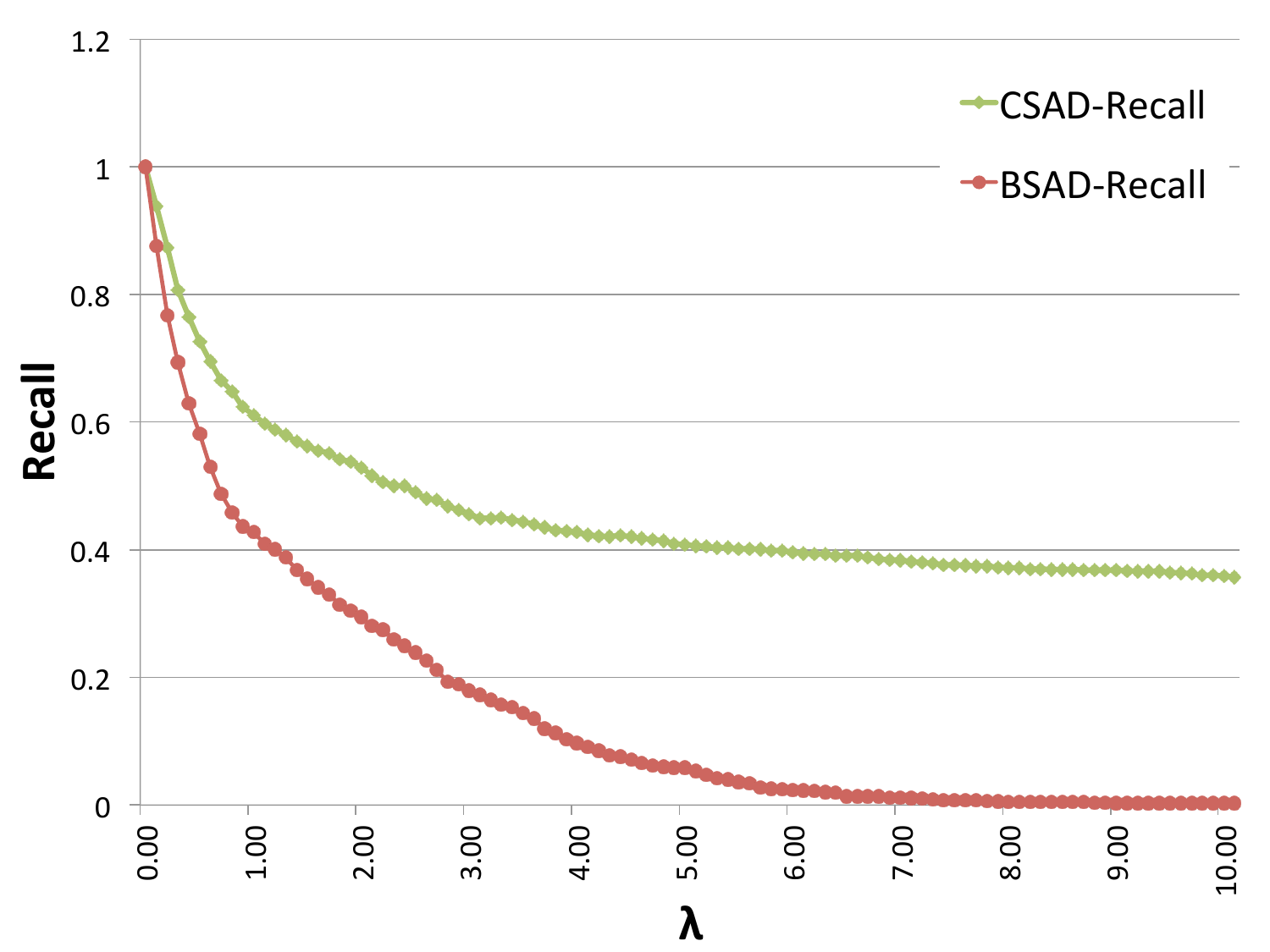}
    \caption{Recall of CSAD and BSAD}
    \label{fig:recall}
\end{figure}

Using this edge annotation, we plot the precision and recall of structural change detection at various values of the lasso regularization parameter $\lambda$ in figures (\ref{fig:precision}) and (\ref{fig:recall}) respectively. With increasing regularization, fewer edges are detected as more entries in the estimated precision matrix are driven to zero. Hence, the precision improves for CSAD.

For BSAD, the precision improves until $\lambda \approx 5.0$, and then deteriorates sharply. This phase change in precision happens precisely because BSAD is non-contrastive. Until $\lambda \approx 5.0$, the structural change is captured correctly as signal, albeit poorly compared to CSAD. After $\lambda \approx 5.0$, the regularization is too strong and the structural change signal is lost as noise in comparison to the effect of the background precision matrix. In CSAD, this does not happen because we estimate the background precision matrix separately before anomaly detection and the second step is dedicated to estimating only the structural change from a learned background precision matrix.

As regularization increases, the recall of both CSAD and BSAD decreases as fewer edges are detected. CSAD provides better recall than BSAD at all values of $\lambda$.

\section{Conclusions and Future Work}

We proposed a method to detect structural changes in GGMs given datapoints from the background GGM and a structurally different foreground GGM. We evaluated the method on a network of 100 nodes and found promising improvements on both precision and recall of structural changes. One direction of future work is to test for statistical significance of detected changes using a statistical test or a scoring mechanism \cite{maurya2016semantic}. Another useful direction of generalization is to identify model misfit of a single GGM and use a mixture of GGMs to model background data as the observed variables might come from a mixture of Gaussian MRFs instead of a single Gaussian MRF.

\section*{Acknowledgments}
We would like to thank Nicole Rafidi for insightful criticism on this project.

\nocite{langley00}

\bibliography{report}
\bibliographystyle{icml2014}


\end{document}